\documentclass[letterpaper, 10 pt, conference]{ieeeconf}  

\IEEEoverridecommandlockouts                              

\usepackage{bm}
\usepackage{times}
\usepackage{epsfig}
\usepackage{amssymb}
\usepackage{algpseudocode}
\usepackage{algorithm}
\usepackage{amsmath}
\usepackage{subfigure}
\usepackage{graphicx}
\usepackage{multirow}
\usepackage{multicol}
\usepackage{caption}
\usepackage{gensymb}
\usepackage{hyperref}
\hypersetup{colorlinks}
\usepackage{mathtools}
\usepackage{booktabs}
\usepackage{pifont}
\usepackage{xspace}

\newcommand{\etal}{\textit{et al.}}

\newcommand{\trsp}{{\scriptscriptstyle\top}}

\title{\LARGE \bf
\textcolor{black}{Representing Robot Geometry as Distance Fields:\\Applications to Whole-body Manipulation}
}

\author{Yiming Li\textsuperscript{1, 2}, Yan Zhang\textsuperscript{1,2}, Amirreza Razmjoo\textsuperscript{1,2}, and Sylvain Calinon\textsuperscript{1,2}
\thanks{$^{1}$ Idiap Research Institute, Switzerland}
\thanks{$^{2}$ Ecole Polytechnique Fédérale de Lausanne (EPFL), Switzerland}
\thanks{Email: \texttt{\{name.surname\}@idiap.ch}}
\thanks{This work was supported 
by the China Scholarship Council (No. 202204910113), 
by the State Secretariat for Education, Research and Innovation in Switzerland for participation in the European Commission's Horizon Europe Program through the INTELLIMAN project (\url{https://intelliman-project.eu/}, HORIZON-CL4-Digital-Emerging Grant 101070136) and the SESTOSENSO project (\url{http://sestosenso.eu/},
HORIZON-CL4-Digital-Emerging Grant 101070310). We also acknowledge support from the SWITCH project  (\url{https://switch-project.github.io/}), funded by the Swiss National Science Foundation.}
}

\begin{document}

\maketitle
\thispagestyle{empty}
\pagestyle{empty}


\begin{abstract}
In this work, we propose a novel approach to represent robot geometry as distance fields (RDF) that extends the principle of signed distance fields (SDFs) to articulated kinematic chains. Our method employs a combination of Bernstein polynomials to encode the signed distance for each robot link with high accuracy and efficiency while ensuring the mathematical continuity and differentiability of SDFs. We further leverage the kinematics chain of the robot to produce the SDF representation in joint space, allowing robust distance queries in arbitrary joint configurations. The proposed RDF representation is differentiable and smooth in both task and joint spaces, enabling its direct integration to optimization problems. 
Additionally, the 0-level set of the robot corresponds to the robot surface, which can be seamlessly integrated into whole-body manipulation tasks. We conduct various experiments in both simulations and with 7-axis Franka Emika robots, comparing against baseline methods, and demonstrating its effectiveness in collision avoidance and whole-body manipulation tasks. Project page: \href{https://sites.google.com/view/lrdf/home}{https://sites.google.com/view/lrdf/home}

\end{abstract}

\section{Introduction}

In robotics, the representation of a robot commonly relies on low-dimensional states, like joint configuration and end-effector poses. However, this low-dimensional representation lacks internal structure details and is insensitive to external factors, limiting the ability to interact with the environment and respond to real-world. To handle this problem, some geometric representations have been proposed, like primitives and meshes, with various applications~\cite{quigley2009ros, zimmermann2022differentiable}. However, they either make simplified assumptions or require significant computational resources to obtain a detailed model.

A natural idea for handling this problem is to encode the geometry of the robot as signed distance fields (SDFs). Several studies in computer vision and graphics have shown the advantages of such a representation~\cite{curless1996volumetric, park2019deepsdf}. Not only does it offer continuous distance information but also exhibits query efficiency. {\color{black}However, despite several robot representations that can be transformed into SDFs, they are either inaccurate (sphere), computationally complex (mesh), or memory-consuming (voxel SDF).} Besides, the SDF representation for articulated objects remains a challenge due to the nonlinear and high dimensionality.

\begin{table}[t]
\begin{center}
\begin{tabular}{cc}
\centering
\multirow{3}{*}[50pt]{\includegraphics[width =0.46\columnwidth]{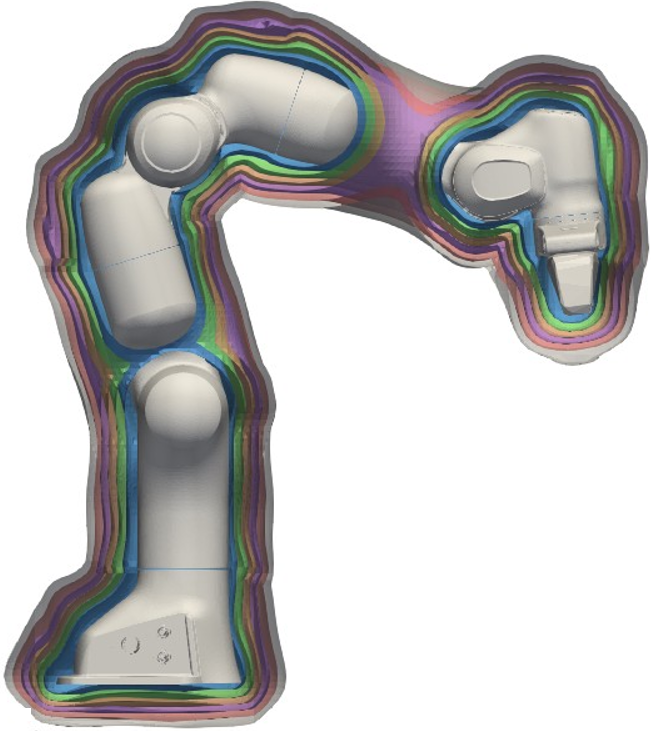}} & 
\includegraphics[width =0.36\columnwidth]{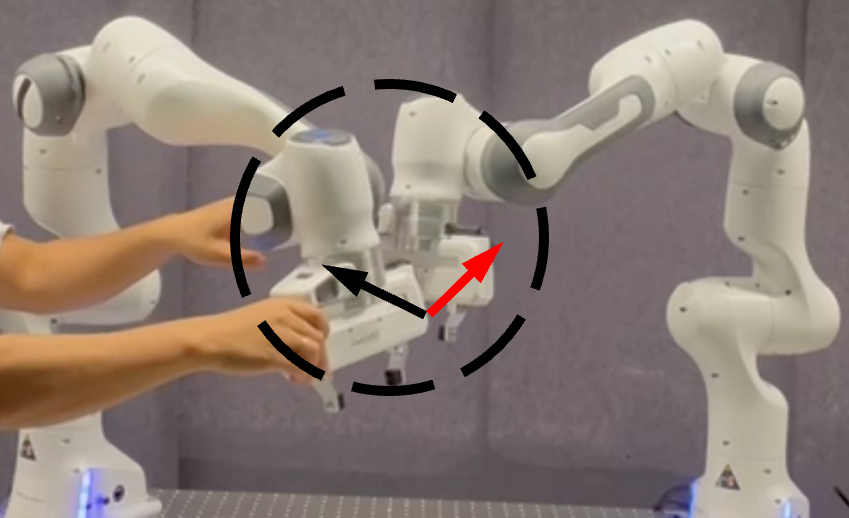} \\
& (b) Collision avoidance \\
& \includegraphics[width =0.36\columnwidth]{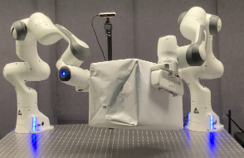}
 \\
(a) Robot SDF & (c) Dual-arm lifting \\  

\end{tabular}
\captionof{figure}{Overview of this work. (a) A precise SDF model of an articulated robot is obtained efficiently by our proposed method. (b) Collision avoidance task based on the SDF representation. (c) Whole-body lifting task with dual-arm.}
\vspace{-5mm}
\label{fig:planned_joints}
\end{center}
\end{table}

{\color{black}Following existing robot representations like spheres and meshes, }we exploit the kinematic structure to represent the distance fields of robots. In contrast to existing methods that encode the robot shape as its joint angle configuration~\cite{liu2022regularized,koptev2022neural}, we adopt a configuration-agnostic approach during the learning phase and utilize the kinematic chain of the robot during the inference phase. This approach simplifies the problem by learning the SDF for each robot link, reducing the dimensionality, and making it robust and reliable for different joint values. During the inference phase, the kinematic information is used to retrieve the SDF values. We utilize Bernstein polynomials as the basis function to represent SDF for each link of the robot for storage and computation efficiency, with facilitated differentiability in task space and joint angle space. 

Representing robot geometry as distance fields (RDF) has multiple advantages. First, it provides a continuous and smooth distance representation, granting easy access to derivatives. This characteristic is particularly well-suited for robot optimization problems such as motion planning and collision avoidance. Further, RDF representation encodes the robot geometry implicitly and decouples from spatial resolutions, enabling whole-body manipulation at any scale without explicitly defining surface points. Finally, RDF allows computationally efficient and precise distance query, which is crucial for various robot applications that require precise perception and quick response, especially in dynamic environments.

We experimentally demonstrate the capabilities of our RDF in three aspects. First, we provide a quantitative comparison of the produced distance fields against other representative methods, showing the advantage of our approach. Then, we conduct collision avoidance experiments to show the real-time control performance. Finally, we present a novel formulation that leverages the RDF representation for manipulation tasks requiring contact, by generalizing the robot's Jacobian matrix from its end-effector to the 0-level set of SDF. We demonstrate the effectiveness of this approach in a dual-arm lifting task, showing how our RDF representation can be seamlessly integrated into first and second-order optimization problems. In summary, our contributions are:
\begin{itemize} 
    \item We propose a simple and flexible structure that leverages Bernstein polynomials to encode SDFs, showing high accuracy and efficiency while ensuring continuity and differentiability.

    \item The proposed SDF representation is further extended to articulated robots by leveraging the kinematics chain, allowing robust interpolation/extrapolation to any joint configuration while keeping the above properties.
    
    \item We demonstrate the effectiveness of our RDF representation in experiments and show how to integrate it into optimization problems for whole-body manipulation tasks without defining any points on the robot surface.

    \vspace{-2mm}
\end{itemize}

\section{Related Work}

\textbf{SDFs for scene/object representation.}
Representing objects or scenes as SDFs is an active research topic in computer vision and graphics due to its query efficiency and the ability to describe complex shapes~\cite{park2019deepsdf, macklin2020local,carr2001reconstruction}. Typically, it is a scalar field defined over a 3D space that assigns signed distance values to points, representing the distance to the surface. The capability of SDFs in modeling object and scenes have shown in versatile applications like mapping~\cite{ortiz2022isdf,izadi2011kinectfusion}, grasping~\cite{breyer2021volumetric,jiang2021synergies} and rearrangement~\cite{danielczuk2021object}. Liu \etal~\cite{liu2021synthesizing} has explored optimizing diverse grasping configurations based on the object SDF. Driess~\etal ~\cite{driess2022learning} proposes to learn kinematic and dynamic models as SDFs for robot manipulation. 

\textbf{SDFs for motion planning.} 
SDFs can also be utilized in optimization problems such as robot motion planning and control~\cite{schulman2014motion,mukadam2018continuous}. CHOMP~\cite{ratliff2009chomp} proposes to use SDF to represent the environment and achieve effective motion planning in trajectory optimization. {\color{black}Schmidt \etal~\cite{schmidt2014dart} uses SDF representation to track articulated objects by exploiting their kinematics structures.}
Sutanto \etal~\cite{sutanto2021learning} extends learning SDFs to approximate generic equality constraint manifolds. Liu~\etal~\cite{liu2022regularized} presents a regularized SDF with neural networks to ensure the smoothness of SDFs at any scale, testing it in collision avoidance and reactive control tasks. Vasilopoulos~\etal~\cite{vasilopoulos2023ramp} sample point on the robot surface and compute their SDF values with GPU acceleration for motion planning. Although representing scenes as SDFs has shown several advantages in robotics tasks, the environment is usually diverse and dynamic, and it is inefficient to obtain SDFs for arbitrary scenes.


\textbf{SDFs for robot geometry representation.} 
An intuitive idea to solve the problem of arbitrary scenes is to represent the robot as an SDF (in addition, or instead of the scene). Koptev~\etal~\cite{koptev2022neural} propose to learn SDFs expressed in joint space with neural networks, allowing query distance values with points and joints as input. Similarly, in \cite{liu2022regularized}, an SDF model is trained with joint angles as input to represent a mobile robot. Michaux~\etal~\cite{michaux2023reachability} introduce a reachability-based SDF representation that can compute the distance between the swept volume of a robot arm and obstacles.

The aforementioned methods learn an SDF model coupled with joint angles, and the performance is limited due to the high dimensional and nonlinear space. In contrast, our approach exploits the kinematic chain which simplifies the problem and leads to better accuracy. A concurrent work~\cite{liu2023collision}, conducted in parallel to our work, also proposes to exploit the kinematic structure with neural networks for collision avoidance. The differences lie in three aspects: 1) Our representation provides interpretable parameters, holds smoothness guarantees and provides an easy way to compute analytic gradients; 2) We further extend this representation to contact-aware manipulation tasks instead of pure collision avoidance; 3) We conduct a detailed comparison with various SDF representations. All approaches exploit parallel computing, implemented with batch operation for both points and joints, consequently showing high computational efficiency. The accuracy of neural network method (one of our baselines) is also better than experimental results in~\cite{liu2023collision}\footnote{Codes are available at \href{https://github.com/yimingli1998/RDF}{https://github.com/yimingli1998/RDF}.}.

\section{Learning Robot Geometry as Distance Fields}
In this section, we present our approach to represent the robot geometry as distance fields. Specifically, we first encode the SDF of each robot link through concatenated Bernstein polynomials and then extend it to the whole body based on the robot kinematic chain.

\subsection{Problem Notations}
Let $\mathcal{R}(\bm{q}) \subset \mathbb{R}^3$ be a robot in the 3D Euclidean space at the configuration $\bm{q}$, and $\partial \mathcal{R}(\bm{q})$ denotes the surface of $\mathcal{R}$. The distance function $f(\bm{p},\bm{q}): \mathbb{R}^3 \rightarrow \mathbb{R}$ is defined with $f(\bm{p},\bm{q}) = \pm d(\bm{p}, \partial \mathcal{R}(\bm{q}))$, where $d(\bm{p}, \partial \mathcal{R}(\bm{q})) = \inf_{\bm{p}' \in \partial \mathcal{R}(\bm{q})}|\bm{p}-\bm{p}'|^2$ denotes the minimum distance between the points $\bm{p} = \{x_1,x_2,x_3\} \in \mathcal{R}^3$ and the robot surface. Signs are assigned to points to guarantee negative values within the robot, positive values outside, and zero at the boundary. The gradient $\nabla f_{\bm p}$ points in the direction of maximum distance increase away from the robot surface. In consequence, the normal $\bm n \in \mathbb{R}^3$ with respect to $\mathcal{R}$ can be defined as $\bm n = \nabla f_{\bm p}$.

\subsection{Kinematic Transformation of SDFs}\label{kinematics_SDF}

Consider a robot with $C$ degrees of freedom and $K$ links, characterized by joint angles $\bm q = \{q_1,q_2,\cdots,q_C\}$ and shapes $\bm \ohm = \{\ohm_1,\ohm_2,\cdots,\ohm_K\}$. The distance field to represent the robot geometry is the minimum of all links SDFs, which can be written as 
\begin{equation}\label{eq1}
    f_{\mathcal{R}} = \min (f_{\ohm_1^b},f_{\ohm_2^b},\cdots,f_{\ohm_K^b} ),
\end{equation}
where $f_{\ohm_k^b}$ is the SDF of link ${\ohm_k}$ in the robot base frame.\footnote{\textcolor{black}{The $\text{min}(\cdot)$ in \eqref{eq1} might cause discontinuous gradient when the closest link changes. A differentiable smooth version of this function can be utilized to avoid this issue.}} 
The SDF value for point $\bm p$ in the robot base frame $f_{\ohm_k^b}$ can be computed through the rigid transformation of SDFs~\cite{driess2022learning}, which involves transforming the query points as
\begin{equation}
    \begin{aligned}
        f_{\ohm_k^b}({\bm p},\bm{q})=f_{\ohm_k}\Big({}^b \mathcal{T}_k^{-1}(\bm q){\bm p}\Big),
    \end{aligned}
\end{equation}
where ${}^b \mathcal{T}_k(\bm q) \in \mathbb{SE}(3)$ denotes a matrix dependent on $\bm q$ that performs the transformation from the frame of the $k$-th link to the base frame of the robot. The computation of these transformation matrices can be achieved using the kinematics chain of the robot, typically represented by Denavit-Hartenberg parameters~\cite{denavit1955kinematic}.

\subsection{Representing SDFs using Bernstein polynomials}
\label{subsec:learn_basis}
\begin{figure}[t]
    \centering
   \includegraphics[width=1.0\linewidth]{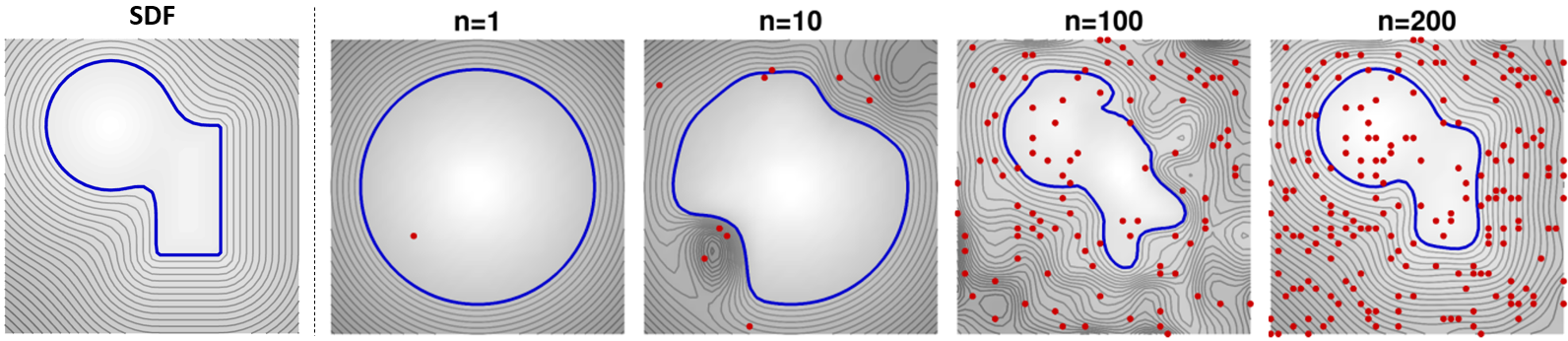}
    \caption{Illustration of iterative learning for a two-dimensional SDF from samples at different locations. The weights are initialized to resemble a circular object. Red points are sequentially sampled for weight updates. The contour of the estimated object shape is depicted by the blue curve (0-level set of the SDF).}
    \vspace{-6mm}
    \label{fig:bf_recursive}
\end{figure}

Basis functions have been widely used in encoding trajectories in robotics, such as in dynamical movement primitives (DMP) \cite{Ijspeert2013} or probabilistic movement primitives (ProMP) \cite{paraschos_probabilistic_2013}, see \cite{Calinon19MM} for a review. They provide a continuous, differentiable, and smooth representation of the trajectory, ensuring the encoded motion appears natural without abrupt changes. This compact parameterization also enables efficient storage and computation while accurately capturing complex motions. 

Drawing inspiration from these studies, we propose the adoption of geometric primitives, a three-dimensional extension of basis functions, to represent the SDF of each link of the robot. By leveraging basis functions with multivariate inputs, we aim to preserve the aforementioned advantages. In this work, we employ Bernstein polynomials, however, other types of basis functions could alternatively be considered, such as Radial Basis Functions (RBF) for infinite differentiability or Fourier basis functions for multiresolution encoding, see \cite{Calinon19MM} for a review.

The SDF $f_{\ohm_k}$ for a point $\bm p^k$ described in the frame of the robot link $\ohm_k$ can be represented as a weighted combination of N basis functions as 
\begin{equation}\label{eq:link_distance}
f_{\ohm_k}(\bm p^k) = \langle\bm \Psi_{\bm p^k},\bm w_k\rangle,
\end{equation}
where $\langle \cdot , \cdot \rangle$ denotes inner product operation, $\bm \Psi_{\bm p^k} \in \mathbb{R}^{1\times N^3}$ is the matrix of basis functions, and $\bm w_k \in \mathbb{R}^{N^3 \times 1}$ is the matrix of superposition weights corresponding to the $k$-th link. We hereby omit the variables $k$ and $\ohm_k$ in order to enhance the readability of the text. We define $\bm \Psi_{\bm p} = \bm \phi(t_1) \otimes \bm \phi(t_2) \otimes \bm \phi(t_3)$ using the Kronecker product $\otimes$, where $t_i = \frac{x_i - x_i^{\text{min}}}{x_i^{\text{max}} -x_i^{\text{min}}}$ is the normalized version of $x_i$, and reshape it to a row vector. $\bm \phi(t) \in \mathbb{R}^N$ is the vector of basis functions whose $n$-th element, for Bernstein polynomials, is given by

\begin{equation}
\phi_n(t) = \binom{N-1}{n}  t^n (1-t)^{N-1-n}, \ \ \ \forall n \in \{ 0,\cdots,N-1\},
\end{equation}
in analytic form, where $t \in [0,1]$ is a normalized location of the point. Consequently, the derivative of the $n$-th basis function can be expressed as
\begin{equation}
\resizebox{.5 \textwidth}{!}{
$     \nabla_t \phi_n(t)= {N-1 \choose n}  (1-t)^{N-n-2}  t^{n-1} \Big( n(1-t)-(N-n-1)t\Big),$
}
\end{equation}
and the derivatives of $\bm \Psi$ are analytically given by
\begin{equation}
\begin{aligned}
        \nabla_{t_1} \bm \Psi = \nabla_{t_1} \bm \phi(t_1) \otimes \bm \phi(t_2) \otimes \bm \phi(t_3), \\
   \nabla_{t_2} \bm \Psi = \bm \phi(t_1) \otimes \nabla_{t_2}\bm \phi(t_2) \otimes \bm \phi(t_3), \\ 
      \nabla_{t_3} \bm \Psi = \bm \phi(t_1) \otimes \bm \phi(t_2) \otimes \nabla_{t_3}\bm \phi(t_3).
\end{aligned}
\end{equation}

For $T$ data points denoted as $\bm P \in \mathbb{R}^{T \times 3}$ and their corresponding distance values denoted as $\bm f \in \mathbb{R}^T$, the weight tensor $\bm{w}$ can be learned through least square regression as $\bm w^{*} = (\bm \Psi^T \bm \Psi)^{-1}\bm \Psi^T \bm{f}$, where $\bm \Psi \in \mathbb{R}^{T \times N^3}$ contains all basis functions and points in a concatenated form. Computing the inverse of large matrices can be computationally expensive and suffer from memory issues. Instead of a batch evaluation, a recursive formulation can be used, providing exactly the same result\textcolor{black}{~\cite{Hager89,Ting10b}}. To do so, we define a new parameter $\bm{B} = (\bm \Psi^{\top} \bm \Psi)^{-1}$ and process the data sequentially by sampling a small batch of points $\{\tilde{\bm P}, \tilde{\bm f}\}$ and updating the learned weights when new data points become available. The whole process is depicted in Algorithm~\ref{algorithm}, and a 2D example is shown in Fig.~\ref{fig:bf_recursive}. After obtaining the optimal weights $\bm w^{*}$, the distance can be decoded efficiently with \eqref{eq:link_distance} during inference. Similarly, this efficiency also extends the gradients, thanks to the analytic form of the polynomial structure\footnote{We refer readers to \href{https://rcfs.ch/}{https://rcfs.ch/} for details about basis functions encoding with multidimensional inputs. }.

\begin{algorithm}[t]
\caption{Recursive learning of superposition weights}
\label{algorithm}
\begin{algorithmic}
\item \textbf{initialize} $\boldsymbol{B}_0= \frac{1}{\lambda} \bm I$, $\bm w =\bm w_0$;
\item \textbf{for} $m \leftarrow 1 \ to \ M$ \textbf{do} 
\item \ \ \ \ Given new mini-batch data points: \{$\tilde{\bm P}, \tilde{\bm f}$\}
\item \ \ \ \ \ Encode points with Bernstein polynomials: $\tilde{\bm \Psi} = {\bm \Psi}(\tilde{\bm P})$
\item \ \ \ \ Compute Kalman gain:\\
\ \ \ \ \ \ $\bm K_m= \boldsymbol{B}_{m-1}\tilde{\bm \Psi}^{\top}(\boldsymbol{I} +\tilde{\bm \Psi} \boldsymbol{B}_{m-1}\tilde{\bm \Psi_m}^{\top})^{-1}$
\item \ \ \ \ Update $\boldsymbol{B}_m$: $\boldsymbol{B}_{m} =\boldsymbol{B}_{m-1} - \bm K_m \tilde{\bm \Psi} \boldsymbol{B}_{m-1} $
\item \ \ \ \ Update $\boldsymbol{w}_m$: $\boldsymbol{w}_m = \boldsymbol{w}_{m-1} + \bm K_m(\tilde{\bm f} - \tilde{\bm \Psi} \bm w_{m-1}) $
\item \textbf{end}
\item return $\bm w^* \leftarrow \boldsymbol{w}_M$
\end{algorithmic} \vspace{-1mm}
\end{algorithm}

\section{Numerical Experiments}

To demonstrate the effectiveness of the proposed method, we conduct several numerical comparisons against baseline methods. 

\textbf{Implement details.} We build the distance field for the Franka Emika Robot with 7 articulations and 9 links (the fingertips of the gripper are ignored). The superposition weights of Bernstein polynomials are separately trained for each robot link. Specifically, we assume a cubic volume around each link to sample training data. The positions of points inside the volume are normalized to $[0,1]$ and points outside the volume are projected onto the boundary and the distance is approximated by summing the distances from the projected point to the boundary. All operations are implemented with the batch operation and run on an Nvidia GeForce RTX3060 GPU. The training data is generated following DeepSDF~\cite{park2019deepsdf}\footnote{we use the library \href{https://github.com/marian42/mesh_to_sdf}{mesh\_to\_sdf} for implementation.}.

\textbf{Effectiveness of basis functions in encoding SDFs.} 
We first compare the proposed Bernstein Polynomial (BP) method with two other representative state-of-the-art approaches: a volumetric-based method, TT-SVD~\cite{boyko2020tt} which utilizes tensor decomposition to compress voxelized SDFs, and a neural network (NN) based method~\cite{park2019deepsdf,liu2023collision}. We evaluate the Chamfer Distance (CD)~\cite{park2019deepsdf}, inference time and model size. For the TT-SVD method, we set the maximum rank $R$ to 40. For the neural network, we found it could not represent the SDF well with the same number of data points we used ($2.56\times10^5$), so we additionally trained it with 10 times more data for a more detailed comparison. We report the result of our lightweight model (with 8 basis functions) and precise model (with 24 basis functions) in Table~\ref{tab1}. Our approach shows competitive accuracy and efficiency with a more compact structure. Although TT-SVD shows a lower mean CD, it exhibits a higher max CD, indicating sensitivity to high-frequency data. Besides, it represents discrete SDFs while our method is continuous and differentiable. We find BP and NN can encode the shape of the robot links accurately with similar CD. However, our method based on recursive ridge regression shows higher data efficiency. Additionally, our approach offers other benefits. The weights learned by BP correspond to the key points, which directly provide interpretable and controllable parameters. Besides, it also provides a simple and efficient way to compute analytical gradients by directly leveraging the derivatives of the basis functions. The continuity and smoothness of the gradient are also guaranteed by construction.

\begin{table}[t]
\caption{Comparison of baseline methods for representing SDFs}
\resizebox{1.0\linewidth}{!}{%
\begin{tabular}{|c|c|c|c|c|c|}
\hline
\multirow{2}{*}{Methods} &  {CD,} & {CD,} &{Inference} & {Model} \\ 
& {mean (mm)} & {max (mm) } &{Time (ms)} & {size (MB)} \\ 
\hline
TT-SVD  & \textbf{0.22}& 23.3 & - & 3.8\\ \hline
NN ($2.56\times10^5$ points)& 1.86& 32.4 & 0.25 & 2.4\\ 
NN ($2.56\times10^6$ points) & 0.57& 14.4 & 0.25 & 2.4\\ \hline
BP (N=8)& 0.91& 21.8 & \textbf{0.21} & \textbf{0.024}\\ 

BP (N=24) & 0.40& \textbf{12.6} & 0.54 & 0.49\\ \hline
\end{tabular}
}
\label{tab1}
\vspace{-5mm}
\end{table}

\begin{table}[t]
\centering
\caption{Comparison with baselines for the produced distance field. Errors are presented in millimeters (mm). }
\label{tab:sdf_comparision}
\resizebox{1.0\linewidth}{!}{%
\begin{tabular}{|c|cc|cc|cc|cl|}
\hline
\multirow{2}{*}{} & \multicolumn{2}{c|}{Points Near}& \multicolumn{2}{c|}{Points Far} & \multicolumn{2}{c|}{Average} & \multicolumn{2}{c|}{Time } \\ \cline{2-7}
& \multicolumn{1}{c|}{MAE}& RMSE & \multicolumn{1}{c|}{MAE}& RMSE & \multicolumn{1}{c|}{MAE}& RMSE & \multicolumn{2}{c|}{(ms)} \\ \hline

Sphere-based & \multicolumn{1}{c|}{6.45} & 11.3 & \multicolumn{1}{c|}{5.49} & 9.49 & \multicolumn{1}{c|}{5.91} & 10.4 & \multicolumn{2}{c|}{2.2} \\ \hline

\textcolor{black}{Mesh-based (coarse)} & \multicolumn{1}{c|}{\textcolor{black}{13.6}} & \textcolor{black}{19.2} & \multicolumn{1}{c|}{\textcolor{black}{6.57}} & \textcolor{black}{16.8} & \multicolumn{1}{c|}{\textcolor{black}{9.70}} & \textcolor{black}{18.0} & \multicolumn{2}{c|}{\textcolor{black}{2.9}}\\ \hline

\textcolor{black}{Mesh-based (precise)} & \multicolumn{1}{c|}{\textcolor{black}{4.09}} & \textcolor{black}{10.21} & \multicolumn{1}{c|}{\textcolor{black}{1.79}} & \textcolor{black}{11.78} & \multicolumn{1}{c|}{\textcolor{black}{2.82}} & \textcolor{black}{11.2} & \multicolumn{2}{c|}{\textcolor{black}{8.7}} \\ \hline

Neural-JSDF & \multicolumn{1}{c|}{28.2} & 31.6 & \multicolumn{1}{c|}{18.7} & 23.4 & \multicolumn{1}{c|}{23.0} & 27.4 & \multicolumn{2}{c|}{\textbf{0.25}} \\ \hline
NN + K.C. & \multicolumn{1}{c|}{1.74} & \textbf{3.57} & \multicolumn{1}{c|}{1.30} & 2.93 & \multicolumn{1}{c|}{1.50} & 3.24 & \multicolumn{2}{c|}{4.7}\\ \hline

\textcolor{black}{BP (N=8) + K.C.} & \multicolumn{1}{c|}{\textcolor{black}{2.85}} & \textcolor{black}{4.55} & \multicolumn{1}{c|}{\textcolor{black}{2.35}} & \textcolor{black}{3.93} & \multicolumn{1}{c|}{\textcolor{black}{2.57}} & \textcolor{black}{4.22} & \multicolumn{2}{c|}{\textcolor{black}{2.4}}\\ \hline

BP (N=24) + K.C.& \multicolumn{1}{c|}{\textbf{1.71}} & 3.59 & \multicolumn{1}{c|}{\textbf{1.18}} & \textbf{2.87} & \multicolumn{1}{c|}{\textbf{1.41}} & \textbf{3.23} & \multicolumn{2}{c|}{5.8}\\ \hline
\end{tabular}
}
\vspace{-2mm}
\end{table}



\begin{figure}[t]
    \centering
    \includegraphics[width=0.95\linewidth]{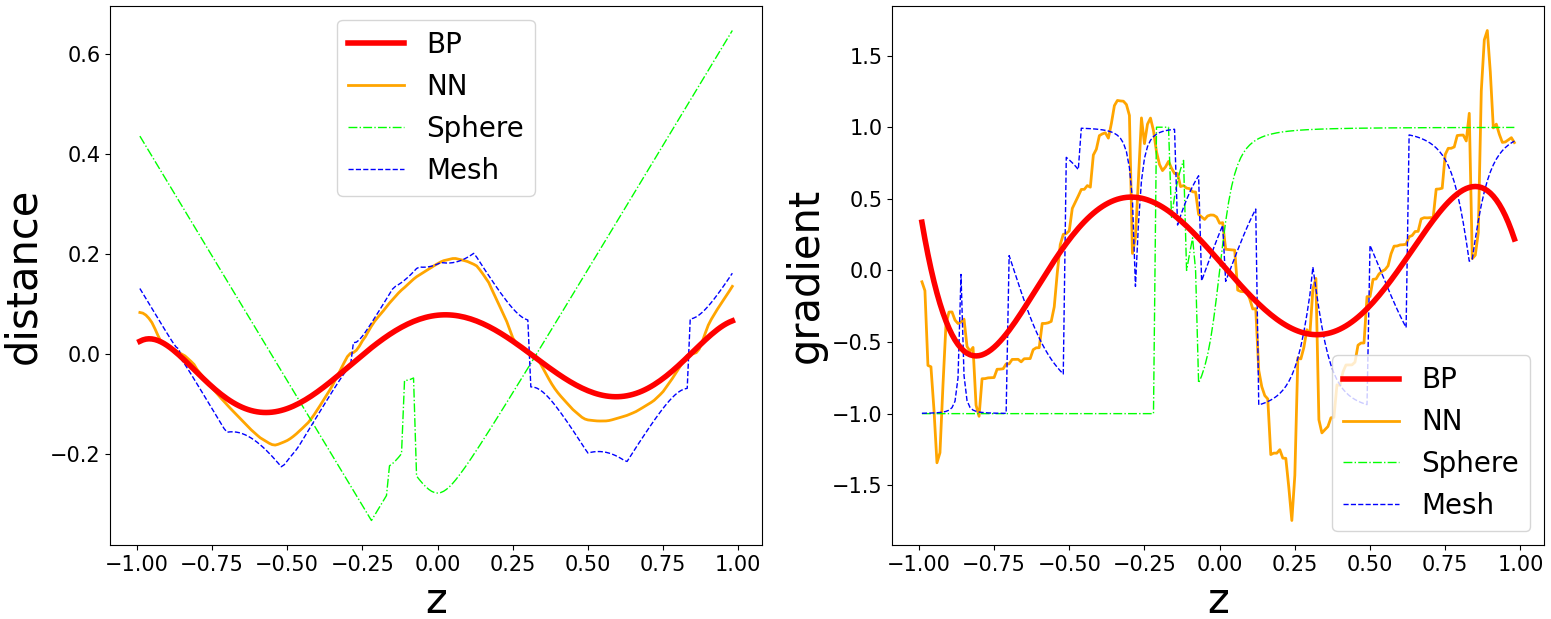}\label{fig:smooth_point} \\
    \includegraphics[width=0.95\linewidth]{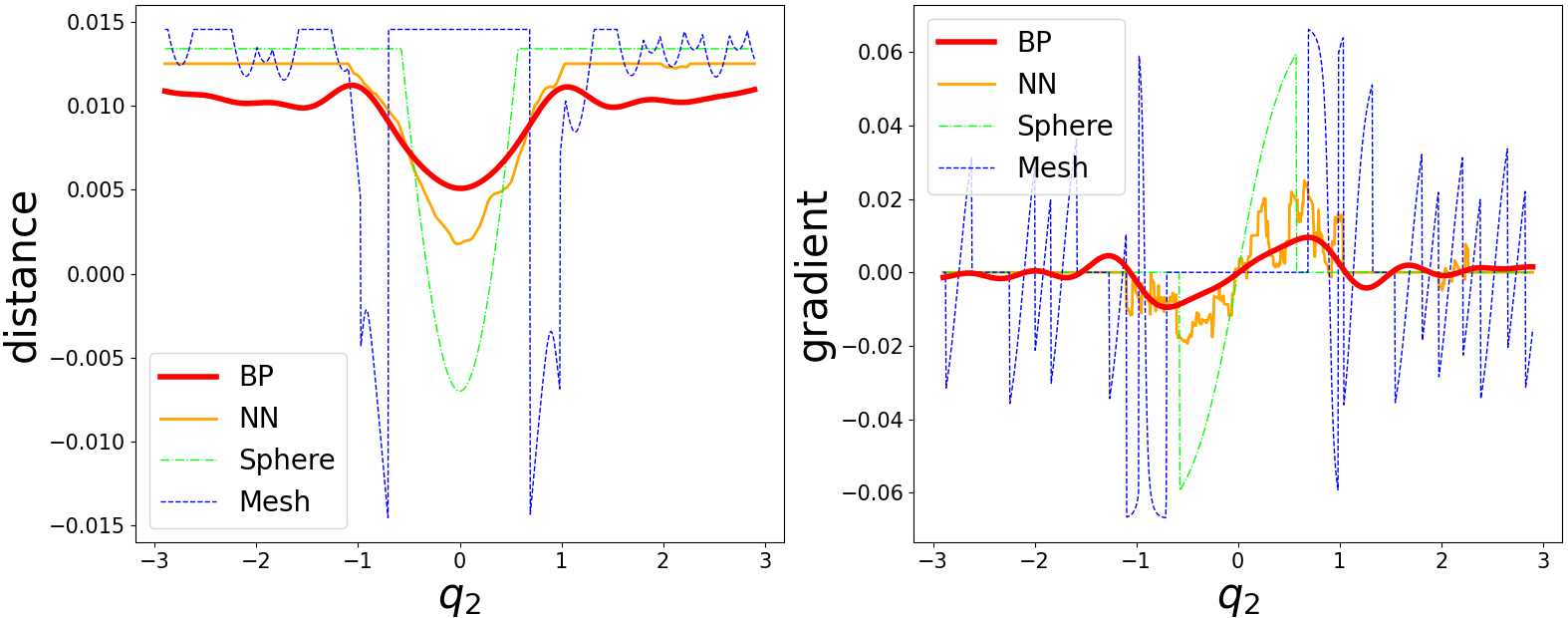}\label{fig:smooth_joint}

    \caption{We show the smoothness of distance and gradient produced by our approach, in both task space (a) and joint space (b), with comparisons to several baselines. The distance and gradient from point $t=[0,0,z]$ to the surface of link5 with a specific joint are shown in (a). The distance and gradient from a specific point to the robot surface at joint $q=[0,q_2,0,0,0,0,0]$ is shown in (b).
    }
    \label{fig:smooth}
    \vspace{-6mm}
\end{figure}


\textbf{Quality of RDF.} We further compare our approach with several common approaches to represent the geometry of the robot. 

Spheres have closed-form signed distance functions and we use 55 spheres to approximate the robot. For meshes, we compute the signed distance by finding the closest vertex and normal, which is another widely used approach. The coarse mesh has 1,249 vertices while the precise mesh has 74,647 vertices. Following Neural-JSDF~\cite{koptev2022neural}, we report the mean absolute error (MAE) and root mean square error (RMSE) for points near the robot surface (within 0.03m) and points far away (over 0.03m) in Table~\ref{tab:sdf_comparision}.


The comparison between Neural-JSDF and other approaches demonstrates the importance of the kinematic chain (K.C.) in modeling accurate RDF. With our non-optimized implementation, the computation time is higher for our method, but it is still at the millisecond level, allowing real-time behavior with high frequency. Although coarse mesh has a more precise shape than spheres, it still fails to represent an accurate distance field, since the choice of closest point and normal estimation are usually inaccurate and noisy. Methods incorporating kinematic chain and SDF (NN $+$ K.C. and BP $+$ K.C.) improve the accuracy compared to primitive-based and mesh-based methods. The average MAE for these methods is about $1\text{mm}$, which is accurate enough for tasks that require establishing contacts with the environment. Figure~\ref{fig:smooth} also shows the distance and gradient produced by several methods, highlighting the smoothness of our approach.

\section{Robot Experiments}

\begin{figure*}[tb]
    \centering
\includegraphics[trim=0.7cm 7.0cm 0.7cm 7.0cm,clip, width=0.98\textwidth]{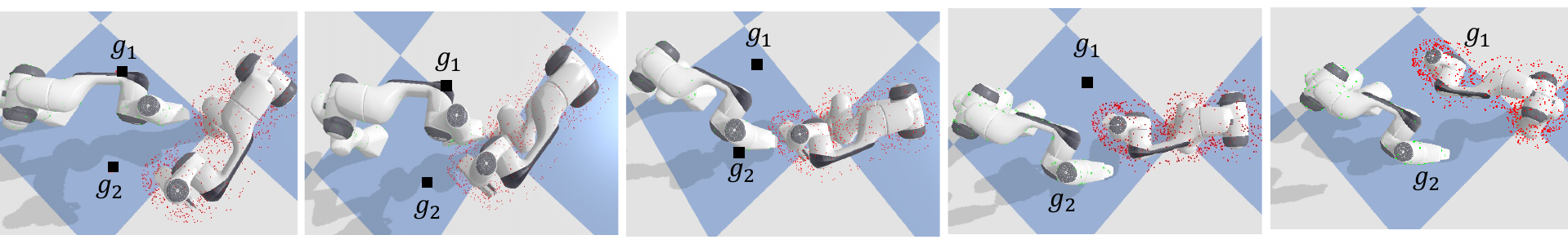}
\caption{Collision avoidance experiment in simulation. $g_1$ and $g_2$ represent the target points. Red points on the right arm are sampled with
the level set $f=0.05$ to represent the safety threshold surface.}
\label{fig:sim_dyn_collision}
\vspace{-4mm}
\end{figure*}

\begin{figure*}[t]
\centering
\includegraphics[trim=0.1cm 6.9cm 0.1cm 6.9cm,clip, width=0.98\textwidth]{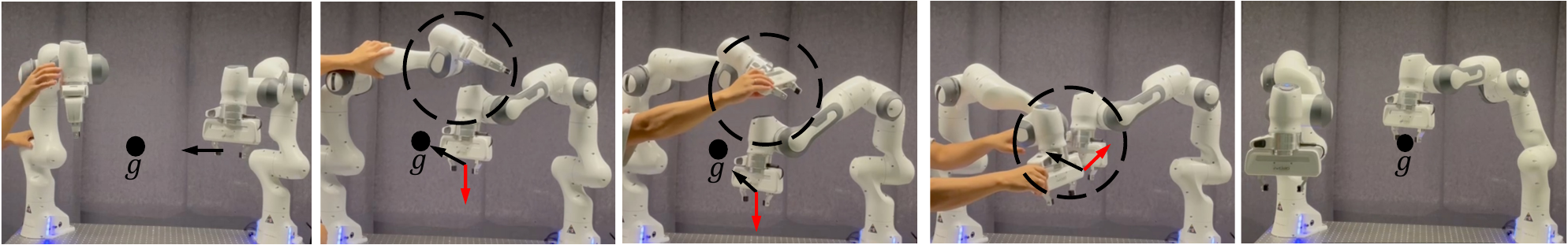}
\caption{Real-world collision avoidance experiment. Here, $g$ is the target point for the right arm. {\color{black} Red/black} arrows show the reaching velocity with/without collision avoidance. Black dashed circles show the potential collision area.}
\vspace{-6mm}
\label{fig:real_dyn_collision}
\end{figure*}
In this section, we illustrate the effectiveness of our RDF representation through two dual-arm robot tasks: 1) \textit{Collision Avoidance}: While a robot arm tries to reach a target, it must avoid colliding with another. 
2) \textit{Dual-arm Lifting}: Two robot arms collaborate to lift a large box that cannot be grasped conventionally. The objective is to plan a pair of joint configurations for both arms such that they can establish contact with the box by exploiting their whole bodies to reach and lift the box.


\subsection{Collision Avoidance}

In this section, we integrate the learned distance fields for collision avoidance, which is crucial in motion planning tasks. Specifically, we exploit an augmented quadratic Programming (QP) algorithm \cite{haviland2023manipulator} to ensure self-collision avoidance between two robot arms during task execution. 

The self-collision avoidance experiments are conducted in both simulation and real-world scenarios. In simulation, the goal for both arms is to reach their respective target position while the right arm should actively avoid collision with the left arm. The real-world experiment is conducted with a reactive controller, where the left arm is manually moved by a human operator in gravity-compensated mode, serving as a dynamic obstacle for the right arm. For both experiments, we randomly sampled 256 points on the surface of the left arm as the input of RDF for the right arm and then used the minimal distance produced for self-collision avoidance.

\begin{table}   
    \centering      \makeatletter\def\@captype{table}\makeatother\caption{Results for collision avoidance in simulation.}
      \label{tab:collision_avoidance}
    \begin{tabular}{|c|c|c|c|} 
     \hline
     Methods & Reaching &  QP no solution & Time cost (ms)\\ 
     \hline
     Mesh-based & 63\% & 37\% & 7.68 \\
     \hline
     NN & 74\% & 23\% &11.34\\ 
     \hline
     Sphere-based & 84\% & \textbf{12\%} & \textbf{4.18}\\ 
     \hline
     BP (N=8) &\textbf{87\%} & \textbf{12\%} &5.99 \\
    \hline
    \end{tabular}
    \vspace{-6mm}
\end{table}

We conducted simulation experiments 100 times, utilizing different initial states for both robot arms, and compared our proposed method with sphere-based, mesh-based and NN-based representations. Results are presented in Table~\ref{tab:collision_avoidance}. The time cost represents the average time for solving the QP problem once. For Neural-JSDF, we observed large distance errors with unsuccessful results. 

As collision avoidance was established as a hard constraint in the QP controller, all methods exhibited collision-free behavior whenever the QP solver converged to a solution. Nevertheless, attributing to the accuracy and smoothness, our method demonstrated the highest success rate $(87\%)$ in reaching, as well as the lowest probability $(12\%)$ of not finding a solution for the QP solver, which also led to a shorter planning time. The left $1\%$ case is that the QP solver found a solution but the two arms blocked each other. The poor performance of mesh-based representation and neural networks indicates the importance of continuous gradient, which makes the optimizer find solutions more easily and more efficiently. Figures~\ref{fig:sim_dyn_collision} and~\ref{fig:real_dyn_collision}
depict the collision avoidance process in simulation and real-world, showing our method enables the robot arm to respond to the environment and avoid collisions (see also accompanying video).

\subsection{Dual-arm Lifting}

\begin{table*}[t]
\begin{center}
\begin{tabular}{ccccc}
\centering
\includegraphics[width =0.37\columnwidth]{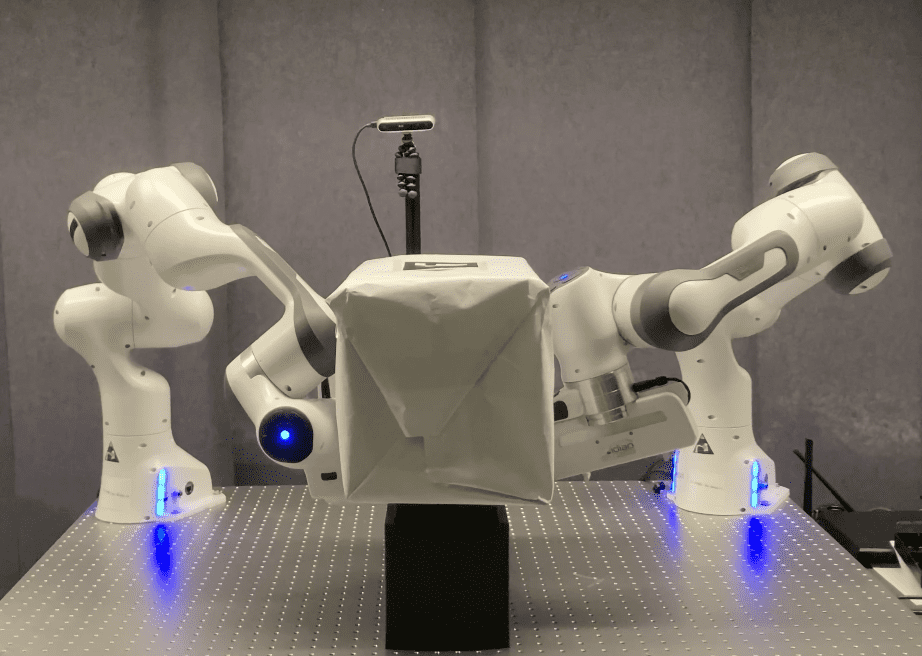}&
 \includegraphics[width =0.37\columnwidth]{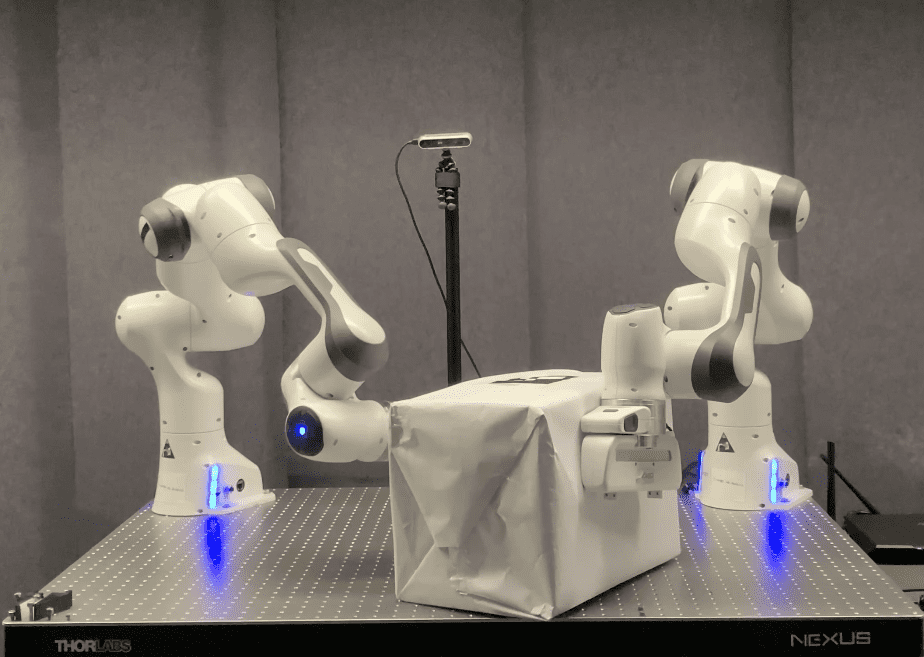}&
 \includegraphics[width =0.37\columnwidth]{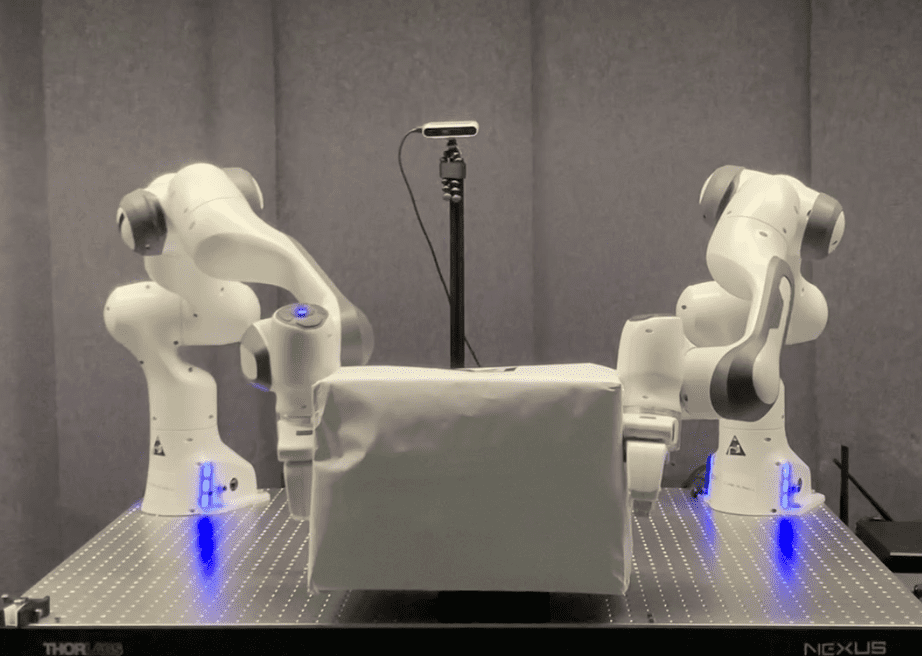}&
 \includegraphics[width =0.37\columnwidth]{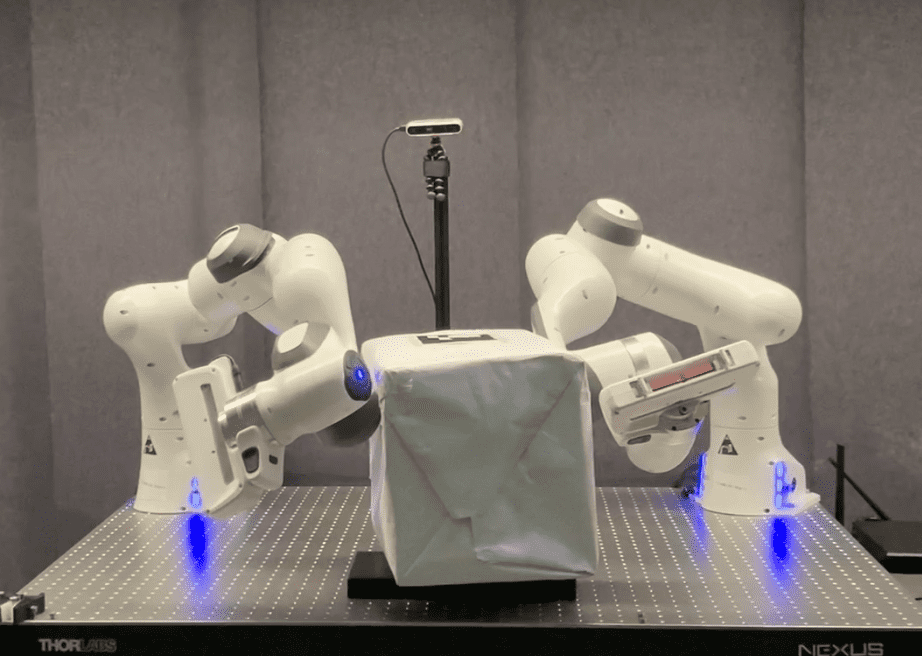}&
 \includegraphics[width =0.37\columnwidth]{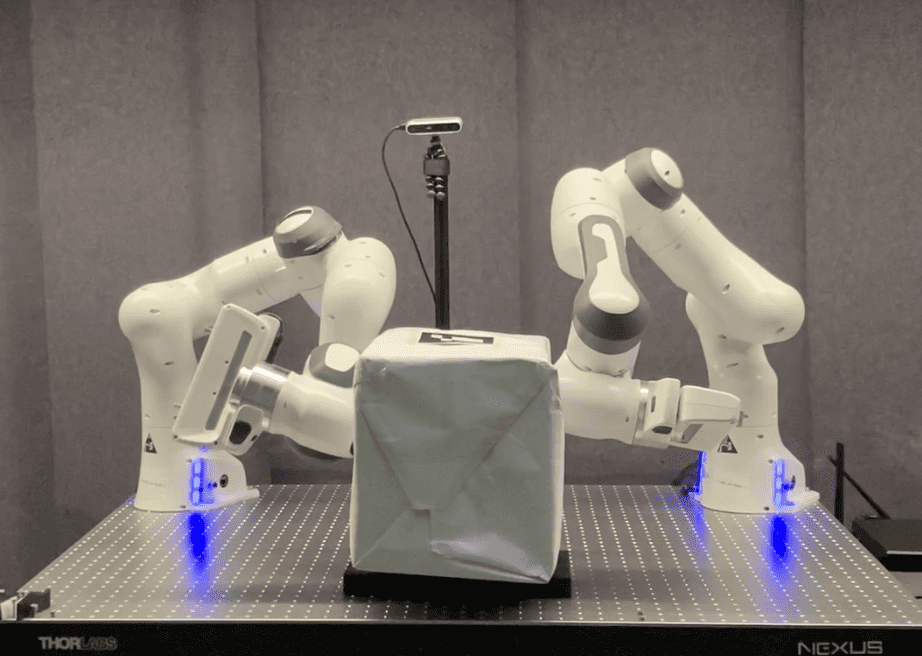}\\

\includegraphics[width =0.37\columnwidth]{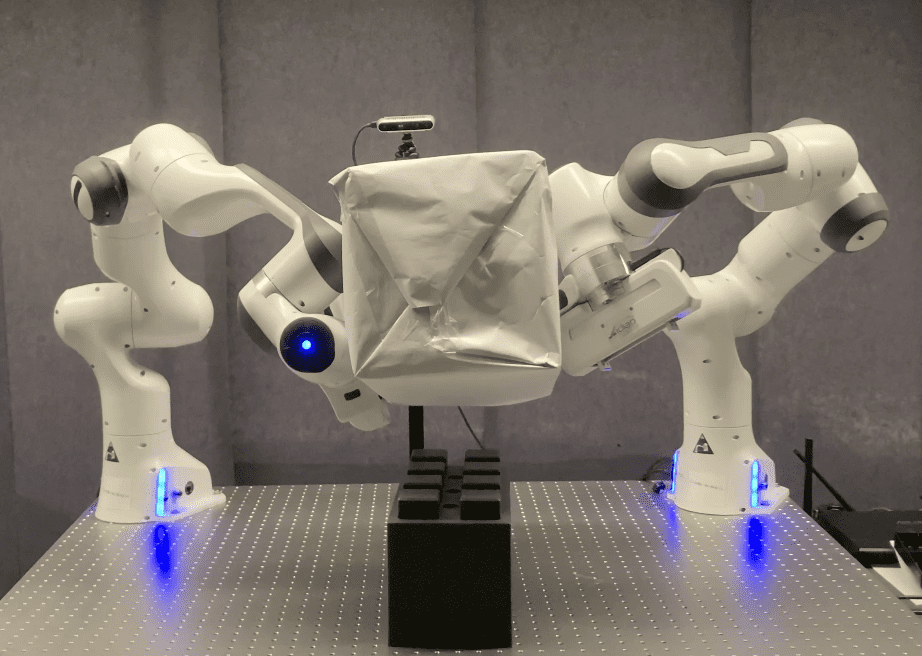}&
 \includegraphics[width =0.37\columnwidth]{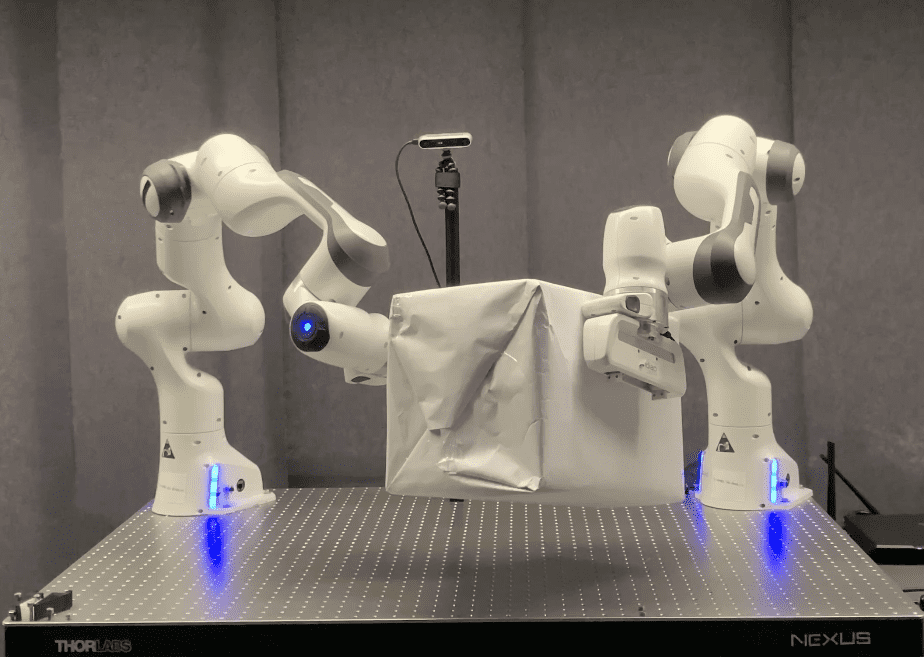}&
 \includegraphics[width =0.37\columnwidth]{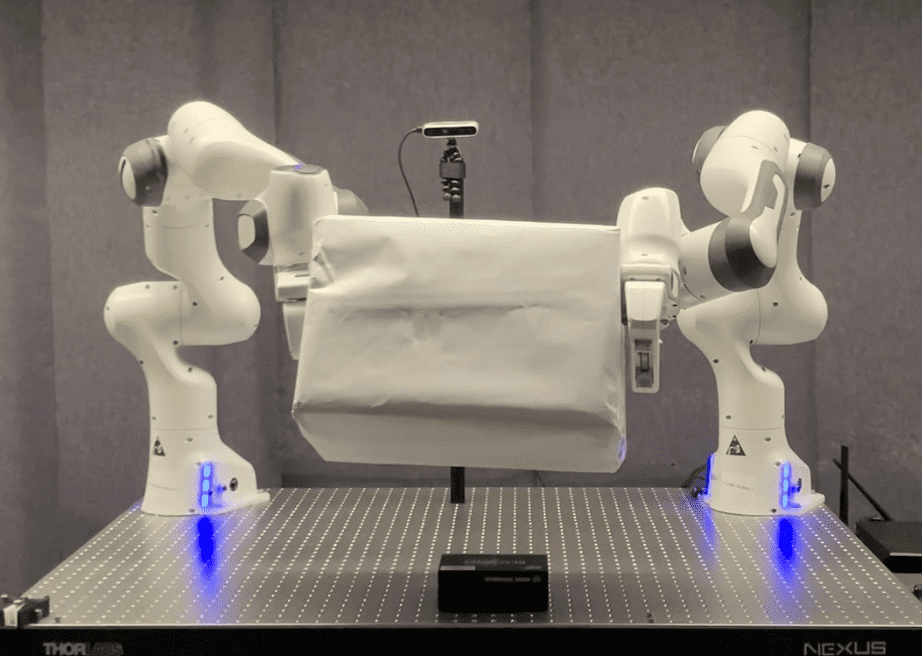}&
 \includegraphics[width =0.37\columnwidth]{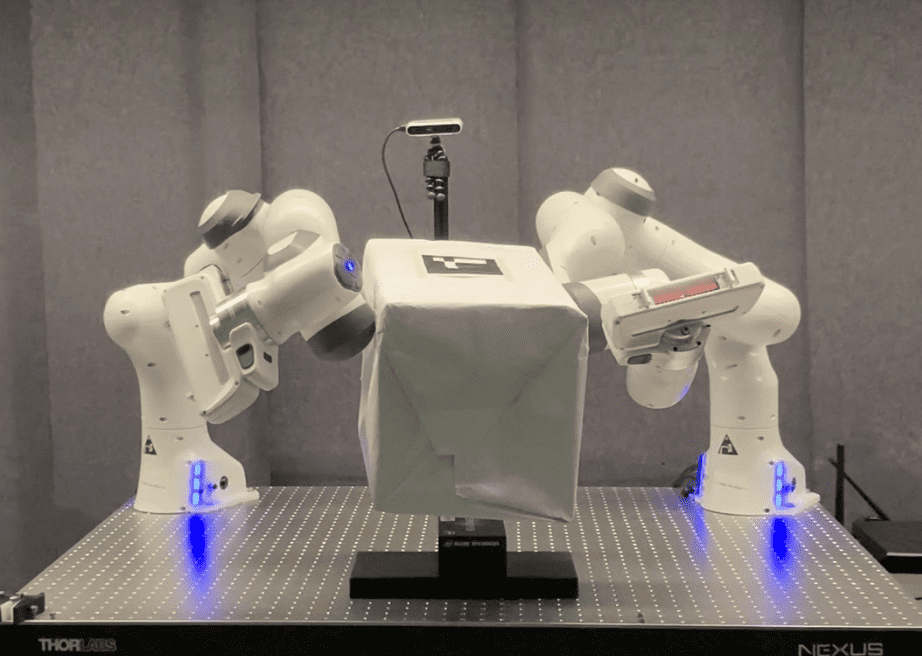}&
 \includegraphics[width =0.37\columnwidth]{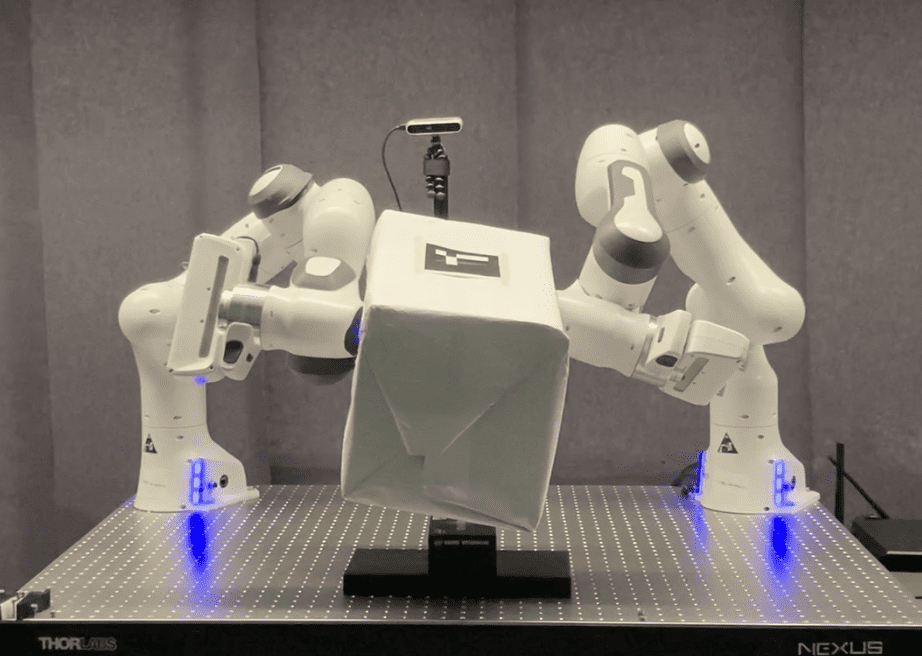}\\
 Exp. 1 & Exp. 2 & Exp. 3 & Exp. 4 & Exp. 5

\end{tabular}
\captionof{figure}{Robot experiments for whole-body dual-arm lifting. \emph{Top row:} the planned joint configurations for grasping the box. \emph{Bottom row:} the final states after lifting.}
\vspace{-8mm}
\label{fig:planned_joints}
\end{center}
\end{table*} 
In this experiment, our focus is on the manipulation of a large box using a dual manipulator, utilizing the whole surface of the last four links of the robot. Our underlying assumption is that the contact points on the object are already predetermined, and the robot has the freedom to establish contacts automatically based on the SDF representation without sampling any surface points on the robot. The task can be viewed as an optimization problem whose quadratic cost function can be defined as $c(\bm{q})= \bm{r}^{\top}\bm{r}$, where $\bm{r} = [\bm r_{r}, \bm r_{c},\bm r_{j}^{\text{max}},\bm r_{j}^{\text{min}},\bm r_{j}^{d}]^{\top} $ is the residual vector consisting of several elements: a reaching residual $\bm{r}_r$ to establish a contact between the robot arms and the object, a penetration residual $\bm{r}_p$ for collision avoidance, a joint distance residual $\bm r_{j}^{d}$ to regularize the solution near the robot’s initial configuration, and joint limit residuals $\bm r_{j}^{\text{max}}$ and $\bm r_{j}^{\text{min}}$ to consider joint angle limits, defined as
\begin{equation}
    \begin{gathered}
        \bm{r}_r = \bm{f}(\bm p_c,\bm q), \quad     \bm{r}_p =  \text{ReLU}(-\bm{f}(\bm p_i,\bm q)), \\
        \bm r_{j}^{\text{max}} = \text{ReLU}(\bm q - \bm{q}_{\text{max}}), \quad
        \bm r_{j}^{\text{min}} = \text{ReLU}(\bm{q}_{\text{min}}-\bm q ), \\
        \bm r_{j}^{d} = \bm q - \bm{q}_{\text{init}}, 
\end{gathered}
\end{equation}
where $\bm{f}(\bm{p},\bm{q}) \in \mathbb{R}^{T}$ represents the spatial distance between points $\bm{p}$ and the robot surface at configuration $\bm{q}$. In this context, $\bm p_c$ represents predefined contact points on the object, while $\bm p_i$ denotes the points uniformly selected within the box for collision avoidance purposes. $\bm{q}_{\text{min}},\bm{q}_{\text{max}}$ are the physical joint limits and $\bm{q}_{\text{init}}$ is the robot initial joint configuration. The optimization is solved using the Gauss-Newton algorithm as 
\begin{equation}
    \begin{aligned}
        \bm q = \bm q - \alpha \bm J^{\dagger}\bm r = \bm q - \alpha(\bm J^{\top}\bm J)^{-1}\bm J^{\top}\bm r,
    \end{aligned}
\end{equation}
where $\bm J = \frac{\partial \bm{r}}{\partial \bm q}$ is the Jacobian matrix and $\alpha$ is a line search parameter. The algorithm is terminated when satisfying criteria $\bm{r}_r^\trsp \bm{r}_r<0.01$, and $\bm{r}_p^\trsp \bm{r}_p<0.01$, and  $q_{\text{min},c} < q_c < q_{\text{max},c}, \; \forall c\in\{1,\ldots,C\}$, and $\sum_{\bm p_c} (1-\langle \text{norm}(\frac{\partial f(\bm p_c,\bm q)} {\partial \bm p_c}),\bm n_c\rangle) < 0.1$ for normal constraints, where $\bm{n}_{\bm{c}}$ is the normal direction on the contact point. We optimize the problem in batch to accelerate the planning procedure with random initialized configurations. Trajectories from initial to goal configurations are interpolated through cubic splines. A joint impedance controller is adopted in conjunction with a smaller desired box size during the planning phase to generate sufficient force at contact points. The lifting action is accomplished by elevating the fourth joint of the robot, which is positioned immediately before the potential contact links. 
 
{\color{black} We report the success rate among 50 planned joint configurations for both arms and the average planning time in Table~\ref{tab:lifting}. A configuration is considered successful if it respects all the termination conditions with the exact robot model. Since batch optimization is usually accompanied by a larger memory overhead, we only selected the sphere-based method and our lightweight model for comparison. Our method shows significant improvement in terms of both success rate and computation time, which is attributed to the more accurate robot model compared to the sphere-based representation.}

\begin{table}
    \centering      \makeatletter\def\@captype{table}\makeatother\caption{Results for dual-arm lifting task.}
\label{tab:lifting}
    {\color{black}\begin{tabular}{|c|c|c|}
     \hline
     Methods & Sphere-based & BP (N=8)\\ 
     \hline
     Success Rate & 36\% & \textbf{77\%}\\
     \hline
     Time (per valid configuration)  & 0.98 &\textbf{0.46}\\ 
    \hline
    \end{tabular}
    }
    \vspace{-6mm}
\end{table} 

The experimental results of the real robot implementation are presented in Fig.~\ref{fig:planned_joints}. In Experiments 1-3, the robot exhibits the capability to use its last four links to contact the object. These experiments provide empirical evidence of the generalization capability of the method across various poses. In experiments 4 and 5, the robot is constrained to utilize specific links for contacts. Specifically, the contact is limited to the sixth link in experiment 4, while in experiment 5, it is restricted to the seventh link. This restriction narrows down the valid solutions, requiring the robot to adapt its approach accordingly. The optimization problem is still able to find appropriate solutions. It can be attributed to the infinite resolution of the robot arm and the smooth representation provided by the distance field, which enables the optimization algorithm to navigate the constrained search space more effectively, leading to successful solutions even in scenarios with limited contact options.



\section{Conclusion}
\label{sec:conclusion}

In this paper, we proposed a novel approach to represent the geometry of a robot as distance fields. We leveraged the kinematic structure of the robot to generalize configuration-agnostic signed distance functions that remain valid for arbitrary robot configurations, which enables more effective learning and more accurate inference of distance fields. The SDF for each link of the robot is represented by a combination of piecewise multivariate polynomials, ensuring interpretability, compactness and smoothness while remaining competitive in terms of efficiency and accuracy. The approach provides analytic derivatives that can directly be used for gradient-based (or higher-order) optimization techniques. Experiments in collision avoidance have shown the effectiveness of our representation. Furthermore, we have demonstrated how to integrate this representation into whole-body manipulation tasks, by defining cost functions based on the SDF of the robot. 

There are some limitations that should be acknowledged. First, the capability of basis functions to highly complex shapes has not been thoroughly investigated. Secondly, we simplified the lifting task by planning joint configurations, without considering the dynamic model. Finally, the representation could be further applied to other complex manipulation tasks, such as pushing and pivoting, by estimating the interaction forces between SDFs and formulating it as an optimization problem. We plan to explore this research direction in future work.

\bibliographystyle{IEEEtran}
\bibliography{main}

\end{document}